\documentclass{article}

\PassOptionsToPackage{numbers, compress}{natbib}

\usepackage[final]{neurips_2022}

\usepackage[utf8]{inputenc} 
\usepackage[T1]{fontenc}    
\usepackage{hyperref}       
\usepackage{url}            
\usepackage{booktabs}       
\usepackage{amsfonts}       
\usepackage{nicefrac}       
\usepackage{microtype}      
\usepackage{xcolor}         

\usepackage{microtype}
\usepackage{arabtex}
\usepackage{utf8}
\setcode{utf8}
 \usepackage{booktabs}
 \usepackage{graphicx}
\usepackage{multicol}
\title{DziriBERT: a Pre-trained Language Model \\ for the Algerian Dialect}

\author{%
  Amine Abdaoui \\
  Oracle \thanks{This work was conducted during the time the author was employed by Geotrend.} \\
  \texttt{amin.abdaoui@oracle.com} \\
\And
 Mohamed Berrimi \\
   University of Ferhat Abbas 1 \\
  Department of computer science\\
  \texttt{mohamed.berrimi@univ-setif.dz} \\
  \AND
  Mourad Oussalah \\
  University of Oulu \\
  Department of CS and Eng.\\
  \texttt{mourad.oussalah@oulu.fi} \\
  \And
  Abdelouahab Moussaoui \\
  University of Ferhat Abbas 1 \\
  Department of computer science\\
  \texttt{Abdelouahab.moussaoui@univ-setif.dz}
}

\begin{document}

\maketitle

\section*{\centering{Abstract }}
Pre-trained transformers are now the de facto models in Natural Language Processing given their state-of-the-art results in many tasks and languages. However, most of the current models have been trained on languages for which large text resources are already available (such as English, French, Arabic, etc.). Therefore, there are still a number of low-resource languages that need more attention from the community. In this paper, we study the Algerian dialect which has several specificities that make the use of Arabic or multilingual models inappropriate. To address this issue, we collected more than one million Algerian tweets, and pre-trained the first Algerian language model: DziriBERT. When compared with existing models, DziriBERT achieves better results, especially when dealing with the Roman script. The obtained results show that pre-training a dedicated model on a small dataset (150 MB) can outperform existing models that have been trained on much more data (hundreds of GB). Finally, our model is publicly available to the community.

\section{Introduction}
Recently, there has been a wide interest in pre-training and fine-tuning large language models using the transformer architecture \cite{mbert,gpt}. 
In contrast to previous word embeddings \cite{wrd2vec, glove}, current language models are trained to generate contextualized embeddings which allow a quality leap in most Natural Language Processing tasks. 
However, most of the current transformers have been pre-trained on languages for which large text resources are already available, such as English \cite{mbert}, French \cite{martin2020camembert} and Italian \cite{italian}. Even multilingual models, such as the mBERT \cite{mbert} and XLM-R \cite{xlmr}, are limited to official languages that have a large web presence. 
Low-resource languages such as African and Arabic dialects received less attention due to the lack of data and their specific and/or complex morphology. For example, the Algerian dialect is spoken by 44 Million people but lacks publicly available datasets.

Indeed, Modern Standard Arabic (MSA) is the most common written language in official documents, books, and newspapers in Algeria. However, the local dialect is very frequent in informal communications, messaging, or in the social media sphere. 
A recent study \cite{arabizi} showed that 74.6\% of the Algerian web-generated content (mostly on Facebook) is conveyed in dialectal Arabic rather than MSA, and 62\% of this content is transcribed in Roman alphabet characters (which is also known as Arabizi). 

The Algerian dialect is mainly inspired from standard Arabic but also from Tamazight\footnote{The original language of the first inhabitants of the region.}, French, Turkish, Spanish, Italian, and English. It has several specificities that make the application of MSA or multilingual models inappropriate. First, it may be written either using Arabic or Roman letters (e.g. Salam \<سلام> (eng: Peace) ). Then, numbers are sometimes used to represent letters that do not exist in the Roman alphabet (e.g. the use of the number 3 to represent the letter \<ع> or the number 7 to represent the letter \<ح >). Finally, despite the influence of the above-cited languages, the Algerian dialect also has its own vocabulary that does not exist in other standard languages.

In this paper, we present a new BERT-like model for the Algerian dialect, named DziriBERT. It has been pre-trained on one Million Algerian tweets.
We evaluate DziriBERT on sentiment, emotion and topic classification datasets. 
The experiments revealed that DziriBERT achieves new state-of-the-art results on several datasets when compared to existing Arabic and multilingual models.

\section{DziriBERT: an Algerian Language Model}
In this section, we describe the collected data and the pre-training settings of DziriBERT.

\subsection{Training Data}
Since there is no available text dataset for the Algerian dialect, we collected 1.2 Million tweets using Twitter API \footnote{https://developer.twitter.com/en/docs} that were posted from major and populated Algerian cities, using a set of popular keywords in the Algerian spoken dialect, such as: {\em ya kho} <eng: my brother>, {\em \<أرواح>} <eng: come>, {\em jibli} <eng: get me>, etc. 
The collected tweets may be written either using Arabic or Latin characters. 
The final dataset after removing all duplicates and entries with less than three tokens contained 1.1 Million tweets (20 Million tokens), which represents almost 150 MB of text data. Then, we performed a light prepossessing on the collected data by (i) replacing all user mentions with \textit{@user}; (ii) all email addresses with \textit{mail@email.com}; and (iii) all hyperlinks with \textit{https://anonymizedlink.com}. Finally, we randomly separate the collected data to a training set (having 1 Million entries) and a test set (having 100 Thousand entries).

The collected dataset is smaller in size when compared to other large scale studies \cite{mbert,arabert}. However, it has been reported that we may need much less data than what we usually use when pre-training language models \citep{martin2020camembert}. The authors have shown that their official model (CamemBERT) trained on 138 GB performs similarly with another version trained only on a sample dataset of 4 GB. Here, we try to push this limit even further.

\subsection{Language Modeling}
DziriBERT uses the same architecture of BERT\textsubscript{Base} (12 encoders, 12 attention heads, and a hidden dimension of 768). First, we train a WordPiece Tokenizer \cite{wu2016google} on our training data with a vocabulary size of 50 Thousand entries. Then, we train our language model using the Masked Language Modeling (MLM) task. Indeed, several studies have shown that the Next Sentence Prediction (NSP) task, originally used in BERT, does not improve the results of downstream tasks \cite{roberta, albert}. 

Since tweets have a short length, we used an MLM probability of 25\% (instead of the usual 15\%). We also set a batch size of 64 due to the limitations of our computational resources. The model has been trained on an AWS g4dn.2xlarge instance\footnote{\url{https://aws.amazon.com/ec2/instance-types/g4/}} with 32 GB of memory and 1 NVIDIA T4 GPU. The training took almost 10 days to complete 50 epochs across the whole training set (around 800k steps). The final model created using PyTorch has been uploaded on the Transformers Hub to facilitate its use \footnote{for anonymity reasons, the link will be added in later versions}.

\section{Evaluation of DziriBERT}
In order to compare DziriBERT with existing models, we have to fine-tune them on downstream tasks. It should also be noted that most of related studies \cite{imen} used either non publicly available dataset or contain only a small part of Algerian dialect, which restricted the scale of potential comparative analysis. 
In this paper, we considered two publicly available corpora that covered both Arabic and Roman scripts: Twifil \cite{moudjari2020algerian} and Narabizi \cite{touileb2021}.

\subsection{Twifil}
\citep{moudjari2020algerian} collected and annotated thousands of Algerian tweets according to the expressed sentiments and emotions. Most of them were written with Arabic letters but there were also many tweets written using the Roman script. The authors shared two publicly available\footnote{\url{https://github.com/kinmokusu/oea_algd}} datasets:
\begin{itemize}
    \item \underline{Twifil sentiment}: 9437 tweets annotated according to 3 polarity classes (positive, negative and neutral);
    \item \underline{Twifil emotion}: 5110 tweets annotated according to the 10 Plutchnik emotion classes \cite{plutchik1984emotions}.
\end{itemize}

\subsection{Narabizi}
The Narabizi corpus, originally published in \cite{seddah2020building}, contains Algerian Arabic sentences written exclusively with the Roman script (Arabizi). In this paper, we use the sentiment and topic classification datasets annotated in \cite{touileb2021}:

\begin{itemize}
    \item \underline{Narabizi sentiment}: 1279 sentences annotated according to 4 sentiment classes (positive, negative, mix and neutral);
    \item \underline{Narabizi topic}: 1279 sentences annotated according to 5 topic classes (sports, societal, politics, religion and none).
\end{itemize}


These four datasets were used to compare DziriBERT with the two most known multilingual transformers (mBERT and XLM-R), and with multiple standard and dialectal Arabic models
(AraBERT, QARiB, CamelBERT and MARBERT). Among the four available versions of CamelBERT, we evaluated the dialectal version (CamelBERT-da) and the one that has been pre-trained on a mix of all datasets (CamelBERT-mix). 

Following 
the work done in \cite{mbert}, we used the final hidden state of the classification token ([CLS]) as a sentence representation followed by one linear layer as a classifier. All models have been fine-tuned for three epochs using the Trainer Class of the Transformers library \cite{transformers} with its default settings. Ten different runs have been conducted for each model on each dataset according to the same 10 seeds that have been randomly generated. Still, the presented results may be reproduced 
using the shared Github repository \footnote{ \url{https://github.com/alger-ia/dziribert}(the generated 10 seeds are listed in the evaluation script)}.

Tables \ref{tab:twifil_dataset}, and ~\ref{tab:narabizi_dataset} present the obtained results on the Twifil and Narabizi datasets. We calculate the accuracy and the macro averaged precision, recall, and F1 score for each model on each dataset.


\begin{table}[]
\centering
\caption{Accuracy and macro averaged Precision, Recall and F1 score obtained on the Twifil datasets}
\label{tab:twifil_dataset}
\begin{tabular}{llcccclccc}
\hline
 &
  \multicolumn{4}{c}{\textbf{Twifil sentiment}} &
   &
  \multicolumn{4}{c}{\textbf{Twifil emotion}}  \\ \cline{1-5} \cline{7-10} 
\textbf{Model} &
  \textbf{Acc.} &
  \textbf{F1.} &
  \textbf{Pre.} &
  \textbf{Rec.} &
   &
  \textbf{Acc.} &
  \textbf{F1.} &
  \textbf{Pre.} &
  \textbf{Rec.} \\ \cline{1-5} \cline{7-10} 
mBERT          & 74.2          & 73.8 & 75.2 & 73.0          &  & 62.0 & 26.0 & 33.3          & 27.0 \\
XLM-R          & 79.9          & 79.5 & 80.9 & 79.1          &  & 64.9 & 26.1 & 26.5          & 28.1 \\
AraBERT        & 73.8          & 73.2 & 74.9 & 72.3          &  & 64.6 & 30.3 & 38.0          & 30.7 \\
QARiB          & 78.8          & 78.2 & 79.0 & 77.9          &  & 68.9 & 39.2 & 42.2          & 38.7 \\
Camel-BERT-da  & 75.2          & 74.6 & 76.0 & 74.0          &  & 66.0 & 34.6 & 38.7          & 34.6 \\
Camel-BERT-mix & 77.7          & 72.2 & 78.6 & 76.7          &  & 69.1 & 38.2 & \textbf{43.8} & 37.5 \\
MARBERT        & \textbf{80.6} & 79.9 & 80.7 & \textbf{79.6} &  & 70.2 & 39.1 & 41.7          & 39.4 \\ \cline{1-5} \cline{7-10} 
DziriBERT &
  80.5 &
  \textbf{80.0} &
  \textbf{81.1} &
  79.5 &
   &
  \textbf{70.4} &
  \textbf{40.1} &
  42.8 &
  \textbf{39.6} \\ \cline{1-5} \cline{7-10} 
\end{tabular}%
\end{table}


\begin{table}[]
\centering
\caption{Accuracy and macro averaged Precision, Recall and F1 score obtained on the Narabizi datasets}
\label{tab:narabizi_dataset}
\begin{tabular}{lcccclcccc}
\hline
 &
  \multicolumn{4}{c}{\textbf{Narabizi sentiment}} & &
  \multicolumn{4}{c}{\textbf{Narabizi topic}}
  \\ \cline{1-5} \cline{7-10} 
\textbf{Model} & \textbf{Acc.} & \textbf{F1.}  & \textbf{Pre}  & \textbf{Rec.} & \textbf{} & \textbf{Acc.} & \textbf{F1.}  & \textbf{Pre}  & \textbf{Rec.} \\ \cline{1-5} \cline{7-10} 
mBERT          & 52.6          & 49.3          & 50.5          & 49.5          &           & 49.3          & 30.8          & 33.8          & 34.1          \\
XLM-R          & 41.9          & 32.2          & 38.1          & 38.3          &           & 43.6          & 21.4          & 19.3          & 27.2          \\
AraBERT        & 49.1          & 46.0          & 47.9          & 47.7          &           & 42.8          & 20.8          & 19.4          & 26.5          \\
QARiB          & 55.0          & 52.9          & 53.7          & 53.4          &           & 45.7          & 29.7          & 29.9          & 32.4          \\
Camel-BERT-da  & 40.9          & 35.5          & 36.0          & 40.1          &           & 43.7          & 21.5          & 20.2          & 27.3          \\
Camel-BERT-mix & 49.4          & 48.3          & 49.4          & 49.6          &           & 47.0          & 27.5          & 25.8          & 31.4          \\
MARBERT        & 58.0          & 55.5          & 56.3          & 55.7          &           & 49.0          & 31.0          & 29.9          & 34.1          \\ \cline{1-5} \cline{7-10} 
DziriBERT      & \textbf{63.5} & \textbf{61.2} & \textbf{62.0} & \textbf{61.4} &           & \textbf{62.8} & \textbf{54.8} & \textbf{64.0} & \textbf{53.2} \\ \cline{1-5} \cline{7-10} 
\end{tabular}%
\end{table}

\section{Discussion}
As shown in Table \ref{tab:twifil_dataset}, DziriBERT and MARBERT \cite{marbert} achieved the best results on the Twifil datasets (which are mainly composed of Arabic script). These two models, which are both pre-trained on tweets, yielded better results than all other multilingual, standard Arabic, and dialectal Arabic models. 
However, DziriBERT yielded much better results on the Narabizi datasets (which are exclusively composed of Roman script)
as shown in Tables \ref{tab:twifil_dataset} and \ref{tab:narabizi_dataset}. MARBERT is again in the second position but the difference with DziriBERT is much more important (+5.5\% in sentiment accuracy and +13.8\% in topic classification). 

An error analysis step revealed that the Twifil datasets contain several entries that are not written in Algerian Arabic. DziriBERT tends to fail more often than MARBERT on documents that are written in standard Arabic or in other Arabic dialects, which may also explain the good results obtained by MARBERT on Twifil. Overall, our experiments have shown that DziriBERT can yield very good results despite the size of its  pre-training dataset. For example, MARBERT has been trained on 128 GB of text (almost x1000 times larger than our pre-training corpus). 
Still, DziriBERT is at least as good as MARBERT on the Algerian dialect and even much better when dealing with Roman characters. 


Furthermore, DziriBERT's vocabulary contains a relatively small number of tokens when compared to the other baselines. Since the embedding layer concentrates most of the model parameters 
\cite{abdaoui}, reducing the number of tokens should have a significant impact on the final model size (which should facilitate its deployment on Public Cloud Platforms). 
Table~\ref{tab:size} presents the vocabulary length, the total number of parameters, and the final size on disk for all models studied here. As expected, even if all models share the same architecture (12 encoders, 12 attention heads, and 768 hidden dimensions), the total number of parameters varies from 110 Million to 278 Million. With its 50k vocabulary, DziriBERT is therefore one of the smallest models studied here.

\begin{table}
\centering
\begin{tabular}{lcccc}

\hline
\textbf{Model} & \textbf{Vocab.} & \textbf{\#Params} & \textbf{Size} \\
 & & \textbf{(Million)} & \textbf{(MB)} \\
\hline
mBERT & 106k & 167 & 672 \\
XLM-R & 250k & 278 & 1147 \\
AraBERT & 64k & 135 & 543 \\
QARiB & 64k & 135 & 543 \\
Camel-BERT-da & 30k & 110 & 439 \\
Camel-BERT-mix & 30k & 110 & 439 \\
MARBERT & 100k & 163 & 654  \\ \hline
DziriBERT & 50k & 124 & 498  \\ \hline

\end{tabular}
\caption{Models comparison according to the vocabulary length, the total number of parameters and the final size on disk.}
\label{tab:size}
\end{table}

\section{Conclusion}
In this paper, we have presented the pre-training and evaluation of DziriBERT, a BERT-based model for the Algerian dialect. Our experiments have shown that this model can achieve good performance on various NLP tasks, even when trained on a relatively small amount of data. In order to support the development of NLP applications for this low-resource dialect, we are making our pre-trained model publicly available. We will also release fine-tuned versions of the model for sentiment analysis, emotion detection, and topic classification.
A natural future work would concern the compilation of more datasets for the Algerian Dialect and the comparison with more recent pre-trained models such as charBERT \cite{djame}.\\ We hope that our work will help to advance the state of the art in NLP for the Algerian dialect and enable researchers and developers to build more effective and useful applications for this language
\bibliography{references.bib}

\end{document}